%% file: main.tex
\documentclass[letterpaper, 10 pt, conference]{ieeeconf}

\IEEEoverridecommandlockouts

\usepackage{etextools}
\usepackage{amssymb}
\usepackage{float}
\usepackage{makeidx}
\usepackage{amsmath}
\usepackage{bbm}
\usepackage{epsfig}
\usepackage{epsf}
\usepackage{psfrag}
\usepackage{verbatim}
\usepackage{color}
\usepackage{multirow}
\usepackage{tabularx}
\usepackage{booktabs}
\usepackage[tight,footnotesize]{subfigure}
\usepackage{array}
\usepackage{soul}
\usepackage{footnote}
\usepackage{cite}
\usepackage{dblfloatfix}

\usepackage{color, colortbl}
\usepackage[colorlinks,bookmarksnumbered,citecolor=orange,urlcolor=orange]{hyperref}
\usepackage{graphicx}
\graphicspath{{Figures}}
\DeclareGraphicsExtensions{.pdf,.png}
\usepackage{bigstrut}
\usepackage[english]{babel}

\usepackage{csquotes}

\usepackage[T1]{fontenc}
\usepackage{algorithm, setspace}
\usepackage{algpseudocode}
\usepackage{url}
\usepackage{multirow}
\usepackage{float}
\usepackage{xcolor}
\usepackage{hyperref}
 \hypersetup{
     colorlinks=true,
     linkcolor=orange,
     filecolor=orange,
     citecolor=orange,      
     urlcolor=orange,
     }

\newcommand{\p}[1]{\smallskip \noindent \textbf{{#1}.}}
\newcommand{\eq}[1]{Equation~(\ref{eq:#1})}
\newcommand{\fig}[1]{Figure~\ref{fig:#1}}
\usepackage{balance}

\title{\LARGE

RILI: Robustly Influencing Latent Intent

}

\author{Sagar Parekh, Soheil Habibian, and Dylan P. Losey
\thanks{The authors are members of the Collaborative Robotics Lab (\href{https://collab.me.vt.edu/}{Collab}), Dept. of Mechanical Engineering, Virginia Tech, Blacksburg, VA 24061.
\newline
{e-mail: \texttt{\{sagarp, habibian, losey\}@vt.edu}}}
}

\begin{document}
\maketitle

\begin{abstract}

When robots interact with human partners, often these partners change their behavior in response to the robot. On the one hand this is challenging because the robot must learn to coordinate with a dynamic partner. But on the other hand --- if the robot understands these dynamics --- it can harness its own behavior, \textit{influence} the human, and guide the team towards effective collaboration. Prior research enables robots to learn to influence other robots or simulated agents. In this paper we extend these learning approaches to now influence \textit{humans}. What makes humans especially hard to influence is that --- not only do humans react to the robot --- but the way a single user reacts to the robot may change over time, and different humans will respond to the same robot behavior in different ways. We therefore propose a \textit{robust} approach that learns to influence changing partner dynamics. Our method first trains with a set of partners across repeated interactions, and learns to predict the current partner's behavior based on the previous states, actions, and rewards. Next, we rapidly adapt to new partners by sampling trajectories the robot learned with the original partners, and then leveraging those existing behaviors to influence the new partner dynamics. We compare our resulting algorithm to state-of-the-art baselines across simulated environments and a user study where the robot and participants collaborate to build towers. We find that our approach outperforms the alternatives, even when the partner follows new or unexpected dynamics. Videos of the user study are available here: \url{https://youtu.be/lYsWM8An18g}

\end{abstract}


\input{intro}
\input{related}

\input{problem}
\input{theory}
\input{simulations}
\input{user-study}

\input{conclusions}


\balance
\bibliographystyle{IEEEtran}
\bibliography{IEEEabrv,bibtex}

\end{document}

%% file: intro.tex
\begin{figure*}[t]
	\begin{center}
		\includegraphics[width=2\columnwidth]{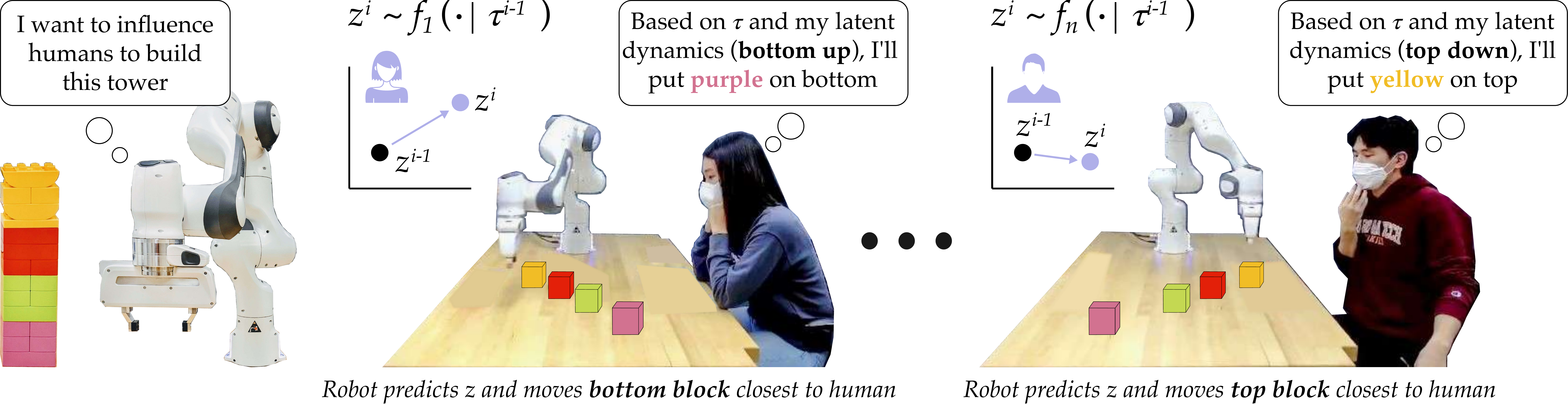}
		\caption{Robot learning to influence human partners to assemble a tower. The tower the human builds \textit{changes} based on how the robot passes the blocks; to coordinate with the current partner, the robot must learn which blocks to move closer or farther from the human. Importantly, different humans respond to the robot in different ways (i.e., have different \textit{latent dynamics}). Our approach tackles this problem by learning to influence changing partners (e.g., humans that build bottom up or top down) and then rapidly adapting the resulting behaviors to new partners (e.g., humans that build from the middle out).}
		\label{fig:front}
	\end{center}
	\vspace{-2em}
\end{figure*}

\section{Introduction}

When robots collaborate with humans in roads or factories, these robots will inevitably affect the actions of their human partners. Imagine working with a robot to build towers (see \fig{front}). During the previous interaction the robot passed you four blocks, and now you must stack these blocks together to form a tower. The way you respond to the robot is personalized. Perhaps you build from the bottom up, using the closest block as the base of your tower; or perhaps you build from the top down, putting the closest block at the top of your tower. The robot has in mind a desired tower that it wants you to build. To reach this tower, the robot must learn how to \textit{influence} your behavior: e.g., if you build from the bottom up, the robot must learn to push the block it wants on the bottom of the tower closest to you. This influential behavior works well so long as you stick with the same rules for responding to the robot --- or, more generally, as long as you maintain the exact same underlying dynamics. But what happens when another person comes along to work with the robot? This new user might have completely different dynamics, so that what the robot learned to influence your behavior no longer works with the current partner.

We consider settings where one \textit{ego agent} (an agent that we control, e.g., a robot) is interacting with one \textit{partner} at a time (e.g., a human). During each interaction the partner has a high-level \textit{strategy} that affects their policy: within our motivating example, the human's strategy is the tower they decide to build (e.g., yellow, red, green, purple). Importantly, the partner's strategy changes based on the previous interaction and the partner's \textit{latent dynamics}. Returning to our motivating example in \fig{front}, perhaps the human's dynamics are to build from the bottom up: so that after the robot pushes a new set of blocks across the table, the human now takes the closest block (e.g., purple) and puts it on the bottom of their tower.

Robots that learn to influence humans within these settings are faced with three key challenges. First, the human's behavior (i.e., their strategy) may change in response to the robot's actions --- for example, the tower the human builds will vary based on the way the robot passes the blocks \cite{habibian2022encouraging}. Second, the robot cannot directly observe the latent dynamics that govern these changes --- e.g., the robot does not know if the human is building from the top down, bottom up, or something else entirely. Finally, different humans will follow different dynamics --- one user might put the closest block on the top of the tower, while another puts the closest block on the bottom. Prior work on multi-agent systems simplifies this problem by assuming that i) the agents are centralized, and we can train the partner alongside the ego agent \cite{hernandez2017survey, foerster2018counterfactual, bowling2002multiagent, cao2018emergent}, ii) the ego agent has access to the partner's strategy and dynamics \cite{tucker2020adversarially, losey2019robots, foerster2017learning}, or iii) the partner always maintains the same latent dynamics, and the ego agent only needs to influence this consistent partner \cite{xie2020learning, wang2021influencing}.

We propose an approach that learns to influence partners without any of these assumptions. Importantly, we recognize that \textit{change is inevitable}: even a single human will have noisy and imperfect dynamics. At first it might seem reasonable to retrain the robot from scratch each time it encounters new dynamics. But this misses out on the connections between partners: even though two partners build the tower in opposite ways, understanding how to push the blocks across the table is still useful for both cases. Rather than learning how to influence every new partner from scratch, or assuming that all partners follow the same latent dynamics:
\begin{center}\vspace{-0.4em}
\textit{Robots can robustly influence new or changing human partners by adapting their latent dynamics while remembering previously learned behaviors.}\vspace{-0.4em}
\end{center}
Robots that apply our approach are able to influence a single human with changing dynamics, or (equivalently) multiple humans with different dynamics. At training time the robot works with a pool of partners over repeated interactions and learns to influence these partners. At test time the robot encounters new dynamics, and transfers the behaviors from the original partners while learning to model these new dynamics. Returning to our example, the robot initially learns to influence human partners that either build from the bottom up or top down; then the robot robustly adapts to and influences new partners that build from the middle out.

Overall, we make the following contributions:

\p{Influencing Changing Partners} We introduce RILI: Robustly Influencing Latent Intent. This algorithm learns to influence partners with changing dynamics by learning multiple latent dynamics models, and then identifying the current partner dynamics from the history of interactions.

\p{Transferring to New Partners} We formalize an extension of RILI that rapidly adapts to new latent dynamics. Our approach i) samples learned behaviors that the ego agent used to influence the original partners, and ii) leverages these behaviors to maximize predicted rewards with new partners.

\p{Testing with Real Human Users} We compare RILI with state-of-the-art baselines across a series of simulated environments. Within our user study the robot learns to influence humans to build a tower: despite the different dynamics that participants used, with RILI the robot influenced humans to build the desired tower in less than $35$ interactions.

%% file: related.tex
\section{Related Work}

\noindent\textbf{Learning alongside Partners.} We focus on settings where one ego agent (e.g., a robot) interacts with a partner (e.g., a human). Here the partner's behavior is \textit{non-stationary} \cite{hernandez2017survey}: returning to our motivating example, the human builds different towers based on how the robot passes blocks. This is challenging because the policy the ego agent learned from previous interactions may fail when the partner \textit{reacts} to that policy. Prior works on multi-agent reinforcement learning address non-stationary behavior by training the agent and partner together \cite{foerster2018counterfactual}, by only training on recent data \cite{bowling2002multiagent}, or by incorporating explicit communication protocols \cite{cao2018emergent}. These centralized approaches are not directly applicable to decentralized human-robot interaction; instead, related research here tries to \textit{model} the partner and predict how their behavior will change \cite{foerster2017learning, ndousse2021emergent}. Our proposed approach also models the partner. But rather than learning the partner's entire low-level policy, we seek to learn a latent representation of their high-level strategy and how this strategy changes.

\p{Influencing Humans} During interaction the ego agent's goal is to maximize its reward. But to maximize this reward, often the ego agent must \textit{influence} the partner it is working with \cite{sadigh2016planning}. Returning to our example: to assemble the correct tower, the robot must get the human to stack the blocks in the right order. Related work on human-robot interaction explores how robots can influence humans using their own behavior, including non-verbal cues \cite{saunderson2019robots}, actions \cite{li2021influencing}, or trajectories \cite{bestick2016implicitly}. Legible motions, where the robot exaggerates its intent, have also been connected with influential behavior \cite{jonnavittula2022communicating}. Most related to our proposed approach are LILI \cite{xie2020learning} and SILI \cite{wang2021influencing}. Both of these works influence partners by learning latent representations of the partner's strategy, and then guiding the partner towards an effective strategy for collaboration. But while \cite{xie2020learning} and \cite{wang2021influencing} have shown promise in \textit{robot-robot} experiments, they have not yet worked with \textit{humans}. Human partners present additional challenges: not only are humans noisy and imperfect, but each human partner may have different reactions, strategies, and dynamics when interacting with the ego agent.

\p{Robust Learning} We therefore turn to prior work that learns \textit{robust} policies for human-robot collaboration. One option is to treat each different partner as a new task \cite{caruana1997multitask} and train the ego agent while interacting with a diverse set of real (or simulated) partners \cite{carroll2019utility}. For instance, training the robot to assemble towers with humans that build from the bottom up, top down, or middle out. Because the robot has already learned to work with multiple different partners, prior work suggests that the robot will be better able to adapt to new users \cite{caruana1997multitask}. Similar to \cite{shih2021critical, lupu2021trajectory}, we break our approach down into two parts: one that is trained across the diverse set of partners, and a second that adapts to the current user.

%% file: problem.tex
\section{Problem Statement}

We consider two-agent settings where we control the ego agent but we cannot control the partner. The ego and and partner \textit{repeatedly interact}. During each interaction the partner has a policy that it follows to select low-level actions, and this policy is parameterized by a high-level \textit{strategy} $z$. In our running example (\fig{front}) the partner's policy is how they reach for, pick up, and assemble the blocks, and their high-level strategy is the tower they want to build. We assume that the latent strategy $z$ is constant within an interaction, but $z$ can change between interactions based on the ego agent's behavior and the partner's \textit{latent dynamics} $f$. For instance, how the robot pushes the blocks across the table influences the next tower that the human chooses to build. The ego agent does not directly observe either $z$ or $f$. Here we formalize this problem setting from the ego agent's perspective, while recognizing that the partner's latent dynamics $f$ can change over time, either because the ego agent is working with a new partner, or because the original partner changes the way they respond to the ego agent.

\p{Hidden Parameter Markov Decision Process} We start by formulating a single interaction. Here $i$ is the index of the current interaction, and the ego agent does not know the partner's current latent strategy $z^i$. We write this interaction as a Hidden Parameter Markov Decision Process (HiP-MDP) \cite{doshi2016hidden}: $\mathcal{M} = \langle \mathcal{S, A, Z, T}, R, H \rangle$. Let $s \in \mathcal{S}$ be the ego's state and let $a \in \mathcal{A}$ be the ego's action. Because this is a two-agent system --- where both the ego agent and partner's behavior affects the interaction --- the transition function $\mathcal{T}(s' \mid s, a, z^i)$ and reward function $R(s, a, z^i)$ may depend on the current latent strategy $z^i \in \mathcal{Z}$. Each interaction has a total of $H$ timesteps, and during the $i$-th interaction the ego agent observes the trajectory $\xi^i = \{ (s_1, a_1), \dots, (s_H, a_H) \}$ of states and actions. Overall, the ego agent experiences $\tau^i = \{ (s_1, a_1, r_1), \dots, (s_H, a_H, r_H) \}$.

\p{Latent Dynamics} Between interactions the partner reacts to the ego agent's behavior and updates their strategy. Recall our running example: because the robot pushed the purple block closest to the partner at the end of interaction $i$, during interaction $i+1$ the partner chooses to put the purple block on the bottom of their tower. The rules that the partner uses to update their strategy --- e.g., building from the bottom up --- are the partner's latent dynamics. We assume these dynamics are Markovian, and only depend on the states, actions, and rewards from the last interaction: $z^{i} \sim f(\, \cdot \mid \tau^{i-1})$.

\p{Playing with Changing Partners} Here we break from prior work and recognize that the partner \textit{does not} always follow the same latent dynamics \cite{xie2020learning, wang2021influencing}. There are two explanations for this (both are equivalent from the ego agent's point of view). First, a single partner could be noisy and shift their dynamics over time; second, the ego agent may encounter multiple partners with different dynamics. For simplicity we refer this as \textit{changing partners}. We assume that the ego agent can practice with $N$ partners, and let the latent dynamics of the $p$-th partner be: $z^{i} \sim f_p(\, \cdot \mid \tau^{i-1})$. But the ego agent will inevitably encounter new partners outside of this training set: returning to our running example, perhaps the robot has trained with partners that build from the bottom up or top down, and then encounters a new partner that builds from the middle out.

\p{Repeated Interactions} Each interaction $i$ from the ego agent's perspective is an HiP-MDP where the partner's latent strategy $z^i$ is unknown. Over the course of the interaction the ego agent observes states, actions, and rewards $\tau^i$. Between interactions i) the partner's latent dynamics $f_p$ may shift and ii) the partner's latent strategy $z^i$ updates according to dynamics $z^{i+1} \sim f_p(\, \cdot \mid \tau^{i})$. Across repeated interactions the ego agent's objective is to \textit{maximize its total reward}. In order to maximize this reward, the ego agent must identify the current partner's latent dynamics (e.g., is the human building from the top down or bottom up?) and then leverage those dynamics to guide the team towards coordinated behavior (e.g., pushing the blocks more or less across the table to get the human to assemble the correct tower).

\begin{figure*}
    \centering
    \includegraphics[width=2\columnwidth]{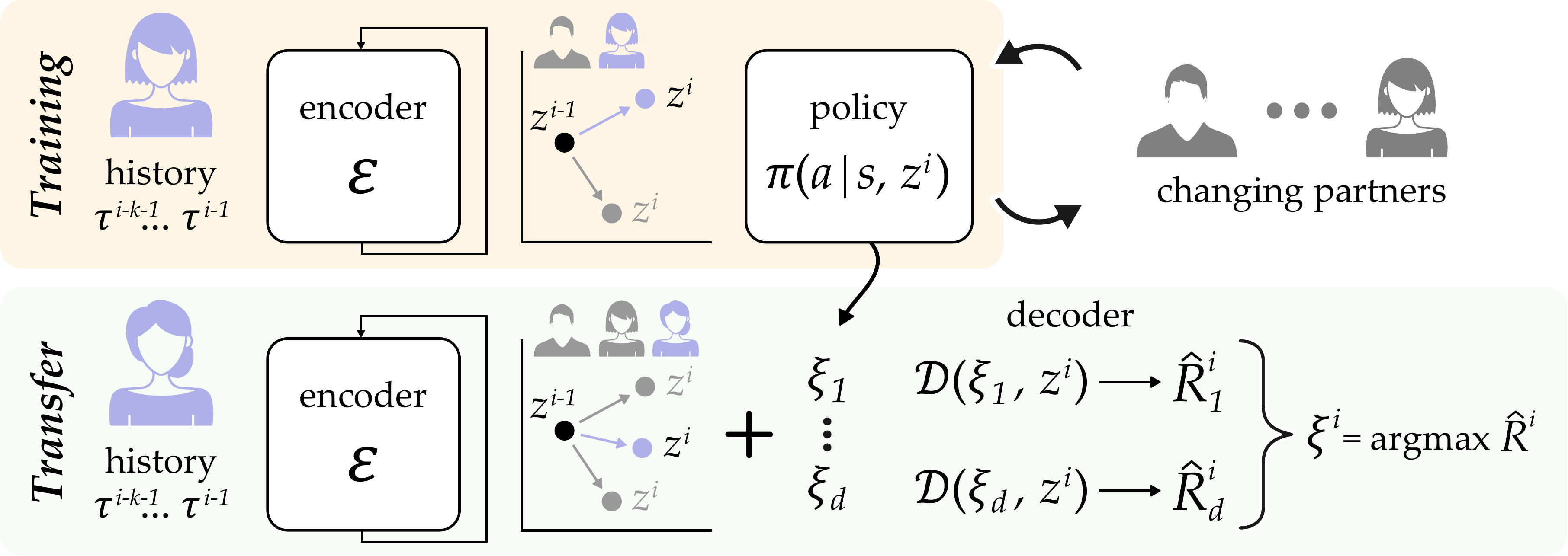}
    \caption{Our proposed RILI algorithm. (Top) During training the ego agent interacts with partners that change latent dynamics without the ego agent's knowledge. Based on the states, actions, and rewards from the last $k$ interactions, the ego agent predicts the partner's next latent strategy $z^i$. We leverage a Gated Recurrent Unit (GRU) encoder $\mathcal{E}$ which identifies the current latent dynamics from the history of interactions. The ego agent then uses model-free reinforcement learning to learn a policy $\pi$ conditioned on the predicted $z^i$. (Bottom) To quickly adapt to new partners with latent dynamics not seen during training, we propose to update the encoder $\mathcal{E}$ and decoder $\mathcal{D}$ while leaving the policy $\pi$ fixed. Here the ego agent samples a set of trajectories $\xi$ from the learned policy, and then executes the trajectory with the highest predicted reward under the learned model for new partner dynamics.}
    \label{fig:main}
    \vspace{-1.5em}
\end{figure*}

%% file: theory.tex
\section{Robustly Influencing Latent Intent}\label{sec:method}

In this section we introduce \textbf{RILI}: \textbf{R}obustly \textbf{I}nfluencing \textbf{L}atent \textbf{I}ntent. We divide our approach into two parts (\fig{main}): first the ego agent trains with changing partners, and learns to influence these partners across repeated interactions. Second the ego agent encounters new partners, and transfers the behaviors it originally learned to now adapt to previously unseen dynamics. We emphasize that the ego agent never observes the partner's strategies or dynamics, and the ego agent does not know when these partners change.

\subsection{Training with Changing Partners} \label{sec:train}

In the first phase of our approach the ego agent learns to influence $N$ changing partners. Each partner has unique latent dynamics $f_p$, $p \in \{1, 2, \ldots, N\}$, and the current partner can switch between interactions. We refer to this as \textit{training} because here we are not focused on rapidly adapting to a specific partner: rather, we want to make the ego agent \textit{robust}, so that it can influence each of these possible partners.

\p{Learning Latent Strategies} Our first challenge is predicting the partner's current latent strategy $z^i$ given the ego agent's recent behavior. Returning to our running example: if the ego agent arranges the blocks in a specific way, what tower will the human assemble? Our insight is that the robot can identify the next latent strategy by recovering underlying patterns from previous interactions. These patterns are specific to the current partner: for instance, if the robot observed high reward in the last few interactions when it pushed the purple block close to the human, perhaps the current user is building from the bottom up. We put this insight into practice by introducing an \textit{encoder} $\mathcal{E}(\tau^{i-1-k}, \ldots, \tau^{i-1})$. This encoder inputs the ego agent's states, actions, and rewards from the last $k$ interactions, and predicts the partner's latent strategy for interaction $i$. At first glance it may seem like this encoder $\mathcal{E}$ only models a single partner $f_p(z \mid \tau^{i-1})$. But the key here is that the encoder reasons over a \textit{history} of $k$ interactions: this history enables the ego agent to \textit{learn multiple latent dynamics} and \textit{identify the current partner}. 
During our experiments we apply a Gated Recurrent Unit (GRU) \cite{cho2014properties} as the encoder $\mathcal{E}$ for learning the dynamics and predicting the strategy.

We have introduced an encoder with the capacity to learn multiple dynamics models --- but how do we train this encoder to actually learn these latent representations? Similar to \cite{xie2020learning, wang2021influencing}, we introduce a \textit{decoder} $\mathcal{D}(\xi^i, z^i)$ that reconstructs the rewards from interaction $i$ given the trajectory of states and actions $\xi^i$ and the predicted latent strategy $z^i$. Intuitively, this means that the latent strategy $z$ must contain enough information such that, when we pair $z$ with $\xi$, the decoder can correctly predict the rewards the ego agent will receive. We train the encoder (with weights $\theta$) and the decoder (with weights $\phi$) across previous interactions:
\begin{equation} \label{eq:autoencoder}
    \min_{\theta, \phi} ~  \sum_{j=k+1}^i\Bigg \| \begin{bmatrix}r_1^j \\ \vdots \\ r_H^j\end{bmatrix} - \mathcal{D}_{\phi}\big(\xi^j, \mathcal{E}_{\theta}(\tau^{j-1-k}, \ldots, \tau^{j-1})\big) \Bigg \|
\end{equation}
where $(r_1^i, \ldots, r_H^i)$ are the ego's rewards at each timestep during interaction $i$. Notice that we train the encoder and decoder only using the low-level states, actions, and rewards the ego agent can observe.

\p{Learning Influential Policies} So far the ego agent has learned a model of its changing partners. Next, we leverage the outputs of this model (i.e., the predicted latent strategies) to guide the ego agent's policy. Remember that the ego agent's ultimate goal is to maximize its total reward. To do this, the ego agent must select behaviors that seamlessly coordinate with the current partner. We therefore condition the robot's policy $\pi$ on our predicted latent strategy: $\pi (a_t \mid s_t, z^i)$. We then train this policy (with weights $\psi$) to maximize the ego agent's rewards across repeated interaction:
\begin{equation}\label{eq:reward}
    \underset{\psi}{\max} \sum_{1}^{N} \sum_{i=1}^\infty \gamma^i \mathbb{E}_{\rho^i_{\psi}} \left[ \sum_{t=1}^H R(s_t, z^i) \right] 
\end{equation}
Here $\gamma \in [0, 1)$ is a discount factor and $\rho^i_\psi$ is the distribution over trajectories $\xi^i$ under policy $\pi_\theta$. Looking at \eq{reward}, the inner summation captures the expected long-term reward for the ego agent when interacting with one specific partner, and the outer summation makes this policy robust by taking into account $N$ changing partners\footnote{In \eq{reward} we assume that each partner is equally likely. When some latent dynamics are more likely than others, the outer summation should be replaced by an expectation over partners.}. \eq{reward} implicitly encourages the ego agent to produce interactions $\tau^{i-1}$ which will influence the current partner towards better latent strategies $z^{i}$ in the next interaction. Returning to our example: the robot must position the blocks during $\tau^{i-1}$ to change the human's strategy $z^i$ and assemble the correct tower.

\subsection{Transferring to New Partners} \label{sec:transfer}

So far RILI has learned to influence a set of $N$ changing partners; next, we \textit{transfer} this learned behavior to rapidly adapt to new and previously unseen dynamics (see \fig{main}). Retraining RILI from scratch (i.e., applying the approach from Section~\ref{sec:train}) is inefficient because i) it can take thousands of interactions before the ego agent coordinates with a new partner, and ii) the retrained RILI policy may fail to influence the $N$ original partners. Here we leverage our insight: behaviors learned to influence the original partners can \textit{still be useful} when interacting with new partners. Consider the robot in our working example --- regardless of the current human's dynamics, the robot still needs to know how to push the blocks across the table. Accordingly, we accelerate adaptation by only retraining \textit{half} of RILI.

We refer to this version of our approach as \textbf{RILI-Transfer}. When we deploy our trained ego agent to interact with new partners, we first \textit{freeze} the learned policy $\pi$. We then resume training the encoder $\mathcal{E}$ and decoder $\mathcal{D}$ using \eq{autoencoder} and the states, actions, and rewards experienced with this new partner. Updating $\mathcal{E}$ and $\mathcal{D}$ enables the ego agent to capture the new partner's latent dynamics. We then select previously learned behaviors that pair well with these new dynamics. More concretely, the ego agent samples\footnote{We use $k$-means clustering across the interaction buffer to find $10 - 80$ diverse and discrete trajectory samples. However, our method is not tied to any specific method for sampling trajectories.} a set of trajectories $(\xi_1, \ldots, \xi_d)$ from its buffer of experiences with the $N$ original partners. Given the predicted latent strategy $z^{i} \sim \mathcal{E}(\tau^{i-1-k}, \ldots, \tau^{i-1})$, the ego agent next applies its decoder to predict the total rewards associated with each trajectory, i.e., $\hat{R}^{i}_j = \sum_{t=1}^H\mathcal{D}(\xi_j, z^{i})$. Here $z^i$ is important: depending on what latent strategy the robot predicts, the same trajectory $\xi$ may have high or low rewards (e.g., pushing the purple block closest to the human works well only if the human builds from the bottom up).  Finally, the ego agent reasons across the predicted reward for each sampled trajectory, and executes the trajectory with the highest reward. We hypothesize that RILI-Transfer will result in fast, robust adaptation to humans because the robot only needs to learn the dynamics model, and does not need to learn a policy for interacting with the environment and partner.

%% file: simulations.tex
\section{Simulations}

\begin{figure*}[t]
	\begin{center}
		\includegraphics[width=1.8\columnwidth]{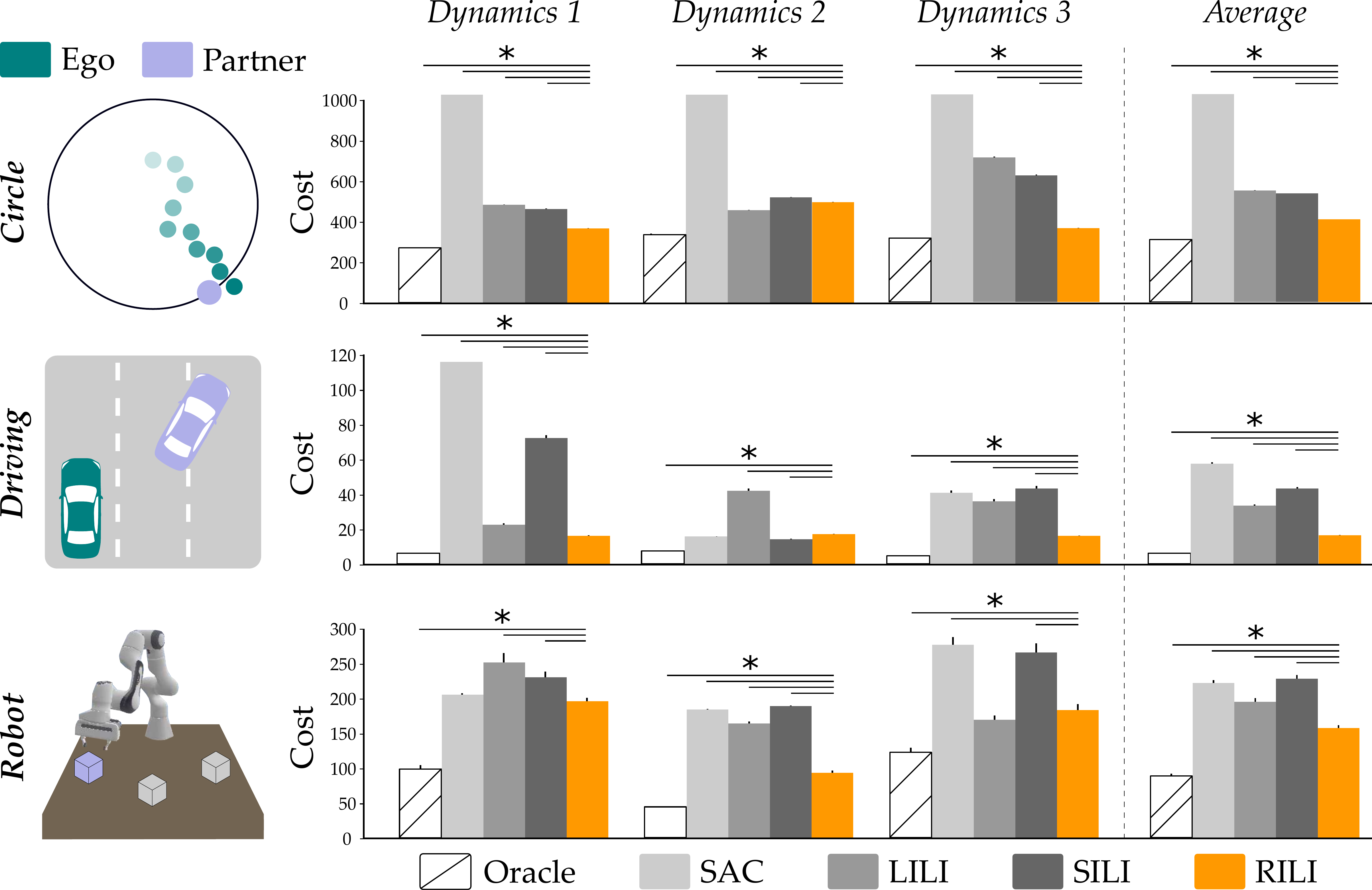}
		\caption{(Left) Simulated environments where the ego agent learns to influence changing partners. The ego agent cannot directly observe the partner's strategy: in \textit{Circle} this strategy is the partner's location, in \textit{Driving} it is the lane the partner merges into, and in \textit{Robot} it is the box the partner wants to pick up. The partner updates their strategy between interactions based on the ego agent's behavior and the partner's latent dynamics. For example, with \textit{Dynamics 1} in \textit{Circle}, the partner moves clockwise if the ego agent ends the interaction outside the circle, and counterclockwise otherwise. (Right) We trained the ego agent as it played with three latent dynamics (i.e., three partners). The dynamics changed randomly between interactions, and the ego agent was never told which dynamics it was interacting with. Here we plot the cost (negative reward) of the final behavior learned by the robot for each of the dynamics, as well as the average cost across all three dynamics. Error bars show standard error about the mean (SEM) and $*$ denotes $p < .001$.}
		\label{fig:sim1}
	\end{center}
	\vspace{-2em}
\end{figure*}

In this section we perform two experiments with simulated human partners. First, we focus on whether embedding a history of interactions enables the ego agent to influence \textit{changing partners} with different dynamics. Next, we evaluate how the ego agent should \textit{transfer} the learned behavior from these original partners to quickly adapt to a new partner. We compare RILI and RILI-Transfer to state-of-the-art baselines across three different simulated environments.

\p{Baselines} We include four different baselines. At one extreme, we consider an ideal case (\textbf{Oracle}) where the ego agent directly observes the partner's latent strategy $z^i$. At the other extreme, we test a learning approach without latent representations (\textbf{SAC}) \cite{haarnoja2018soft}: here the ego agent tries to influence without modeling dynamics or strategies. Between these extremes we implement two related approaches that assume the partner's dynamics are \textit{fixed}. In \textbf{LILI} the ego agent predicts the partner's latent strategy and learns a policy that influences this strategy across repeated interaction \cite{xie2020learning}. \textbf{SILI} builds on \textbf{LILI}, but explicitly encourages the ego agent to guide the partner to a stable, constant latent strategy \cite{wang2021influencing}. 

\subsection{Environments} 

We conducted our simulations in three environments that each have continuous state-action spaces: Circle, Driving, and Robot. These environments are shown in \fig{sim1} with the ego agent in green and the partner in purple. In all environments we defined four different latent dynamics that the partner could follow when responding to the ego agent.

\p{Circle} This environment with two-dimensional states and actions is an instance of a pursuit-evasion game \cite{vidal2002probabilistic} where the ego agent is attempting to reach a hidden partner. The reward is the negative Euclidean distance from the ego agent (a point mass) to the partner (a point on the circle). The partner's latent strategy $z$ is their location on the circle: the ego agent never observes the partner's location, and must learn to find and trap this hidden partner.

Between interactions the partner updates their location using one of four latent dynamics. In \textit{Dynamics~1} the partner moves clockwise if the ego agent ends outside the circle, and moves counterclockwise otherwise \cite{xie2020learning}; in \textit{Dynamics~2} the partner does not move if the ego agent ends outside the circle, and moves counterclockwise otherwise \cite{wang2021influencing}; and in \textit{Dynamics~3} the partner always moves clockwise regardless of the ego agent's behavior. During our first experiment the ego agent practices with these three partners. In the second experiment the ego agent plays with a new partner: this new partner moves counterclockwise if the ego agent ends an interaction outside the circle, and clockwise otherwise.

\p{Driving} Here the ego agent controls its steering angle while attempting to pass a partner in a 2-DoF environment. The ego agent is penalized proportional to the total distance it travels, and it loses $100$ additional points if it collides with the partner car. The partner suddenly changes lanes when the ego agent approaches: the partner's latent strategy $z$ is the lane it merges into, and the ego agent must anticipate this strategy to avoid a collision.

In \textit{Dynamics~1} the partner merges into the lane where the ego agent most recently passed \cite{xie2020learning}; in \textit{Dynamics~2} the partner merges to the far right lane if the ego agent moved to the left early in the interaction \cite{wang2021influencing}; and in \textit{Dynamics~3} the partner cycles through the lanes without reacting to the ego agent. During our second experiment we introduce a new partner. This new partner merges to the far left lane if the ego agent moves to the right early in the interaction; otherwise it moves into the same lane as the ego agent.

\p{Robot} Here the ego agent is a simulated robot arm. There are three goals in the robot's workspace, and the robot is rewarded if it reaches for the same goal as the partner. The robot's action space is its three DoF end-effector velocity, and the partner's latent strategy $z$ is their chosen goal.

In \textit{Dynamics~1} the partner changes their goal to move away from the robot's end-effector; in \textit{Dynamics~2} the partner keeps the same goal if the robot moves to the left of that goal; and in \textit{Dynamics~3} the partner cycles through the goals without responding to the robot. For each of these partners we give a bonus reward to the ego agent if the partner chooses the goal on the right. In our second experiment the ego agent plays with a new partner: this partner maintains the same goal if the robot ends the interaction with its end-effector below a certain height \cite{wang2021influencing}. Additionally, here we give the ego agent a bonus reward each time the partner chooses the middle goal.

\p{Implementation} The input of the GRU encoder $\mathcal{E}$ at interaction $i$ is a history of four interactions $\{\tau^{i-4}, \ldots, \tau^{i-1}\}$, and the output is latent strategy $z^i$ of size $10$. The decoder $\mathcal{D}$ is a multilayer perceptron with a two fully-connected layers of size $64$ each. We use SAC \cite{haarnoja2018soft} to train the ego agent's policy $\pi$: the policy and critic networks are multilayer perceptrons with $2$ fully-connected layers
of size $256$ each. 

\subsection{Playing With Changing Partner Dynamics} \label{sim1}

In Section~\ref{sec:train} we outlined \textbf{RILI}, an approach to robustly learn multiple latent dynamic models and identify the latent dynamics of the current partner. We hypothesized that combining a GRU encoder with a strategy-conditioned policy would enable the robot to learn influential behaviors across changing partners. Here we test \textbf{RILI} --- and the alternatives --- by training the ego agent to play with three changing partners \textit{(Dynamics 1-3)}. Similar to what we would expect in the real world, these partners can change stochastically: after each interaction, there was a $1\%$ chance that the partner would switch. \textbf{Oracle} was given the current parter's latent strategy $z^i$. All other approaches did not know strategy $z^i$, dynamics $f_p$, or even when the partner changed. We trained each ego agent with the same number of interactions. We then took the trained models and evaluated their performance when paired with each partner \textit{(Dynamics 1-3)}.

Our results are shown in \fig{sim1}. Here we report expected cost (i.e., the negative reward) for a single interaction. On the far right column we show \textit{Average}: this is the average cost across \textit{Dynamics 1-3}. Our results suggest that \textbf{RILI} is better able to influence changing partners. For each of the \textit{Dynamics 1-3} and the \textit{Average} we conducted a one-way repeated measures ANOVA with a Sphericity Assumed Correction to test for differences between the four baselines and \textbf{RILI}. Focusing specifically on \textit{Average}, we found that there was a statistically significant difference in the average performance across all three dynamics (\textbf{Circle}: $F(4, 49995) > 30000, p<.001$, \textbf{Driving}: $F(4, 4995) > 1700, p<.001$, and \textbf{Robot}: $F(4, 495) = 216.27, p<.001$). Overall, these results indicate that embedding a history of interactions enables the ego agent to learn and influence the partner dynamics, even when these dynamics change across repeated interaction.

\begin{figure*}[t]
	\begin{center}
		\includegraphics[width=2\columnwidth]{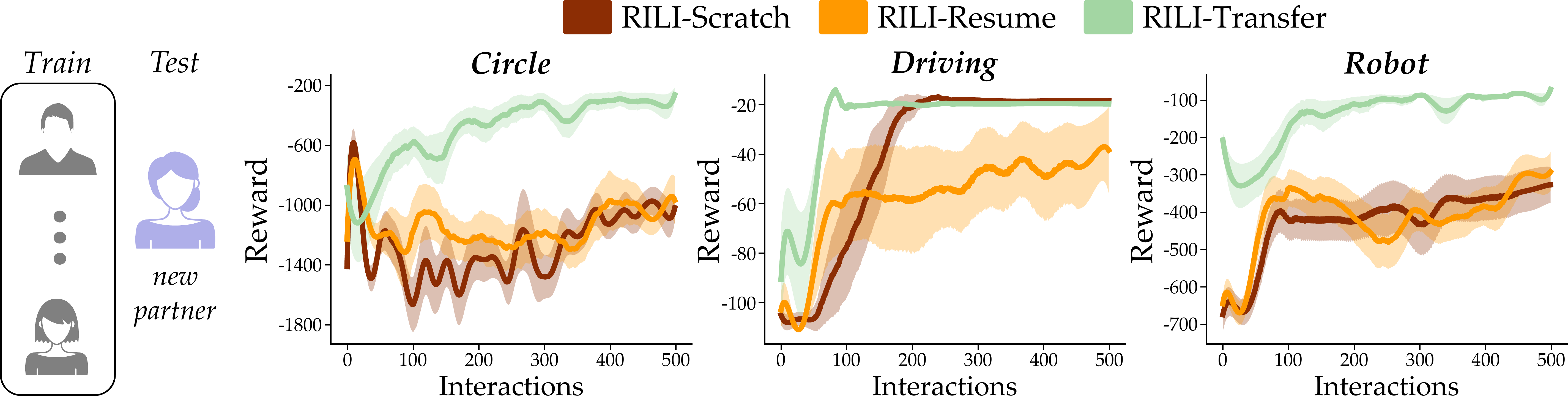}
		\vspace{-0.5em}
		\caption{Follow-up to \fig{sim1}. Here we take what the ego agent learned from playing with \textit{Dynamics 1-3}, and deploy the ego agent with a new partner that uses latent dynamics not seen during training. We compare three versions of our approach. \textbf{RILI-Scratch} throws out what the ego agent previously learned and trains only on the new partner's data. \textbf{RILI-Resume} keeps the learned models from the first three partners, and continues to train these networks with the new partner's data. Finally, \textbf{RILI-Transfer} uses our proposed approach to update only the autoencoder while sampling trajectories from a fixed policy. We find that \textbf{RILI-Transfer} results in rapid adaptation because it re-purposes learned behaviors to rapidly coordinate with the new partner.}
		\label{fig:sim2}
	\end{center}
	\vspace{-2em}
\end{figure*}

\subsection{Playing With New Partner Dynamics} \label{sim2}

Our previous experiment showed that when \textbf{RILI} trains with multiple partners, it learns behaviors to influence these partners and maximize reward. But so far we have not focused on the rate of adaptation, i.e., how many interactions it takes to learn influential behaviors. In practice, this timing is crucial --- human partners may not be willing to spend days working with a confused robot. Accordingly, in this second experiment we take the trained ego agent and encounter a new partner, with dynamics outside of the training set.

Here we test our proposed approach from Section~\ref{sec:transfer}, \textbf{RILI-Transfer}. The intuition behind this algorithm is that the ego agent learns the new partner's dynamics (i.e., retrains the encoder and decoder) while leveraging the behaviors that influenced the original partners (i.e., samples from a fixed policy). We compared this approach to two alternatives: \textbf{RILI-Scratch}, where the robot restarts RILI to learn only from the new partner's data, and \textbf{RILI-Resume}, where the robot keeps the same encoder and policy it has previously learned, and continues to update \textit{both} of these networks.

We tested the performance of each method when playing with a new partner over $500$ repeated interactions. Our results are depicted in \fig{sim2}: we performed $5$ runs, and the shaded regions show the standard deviation across these runs. In all three environments \textbf{RILI-Transfer} rapidly converged to behavior that coordinated with the new partner. For example, in Driving the ego agent had previously learned to pass on the right as part of its strategy for interacting with \textit{Dynamics~3}. When faced with a new partner that merged to the left, the ego agent simply leveraged this existing behavior to coordinate with the new partner and avoid collisions.

%% file: user-study.tex
\section{User Study}

\begin{figure*}
    \centering
    \includegraphics[width=2\columnwidth]{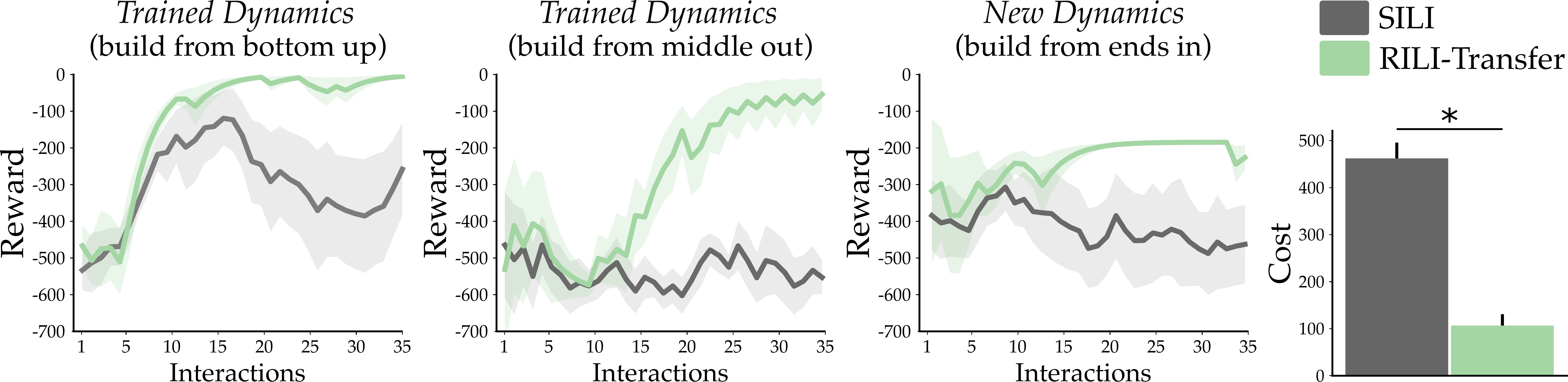}
    \caption{Experimental results from our user study. We recruited $18$ participants to interact with the robot arm and build towers (see \fig{front}). Prior to the experiment we trained the robot to influence four intuitive latent dynamics. During the user study the robot pushed blocks across the table to the human, and then the human assembled these blocks into a tower. Only the robot knew the desired tower. To prevent all participants from adopting the same latent dynamics (i.e., to prevent all users from building from the bottom up), we assigned participants to one of three different latent dynamics. Here we show the performance over time, where each interaction corresponds to building a single tower. We found that \textbf{RILI-Transfer} learned to pass blocks in ways that influenced humans to build the correct tower, and this learning occurred within $35$ interactions. Error bars show SEM and $*$ denotes ($p < .001$).}
    \label{fig:exp}
    \vspace{-1.5em}
\end{figure*}

Our simulations show that agents can leverage RILI to influence changing partners that precisely follow their latent dynamics. But what about humans? Actual users are noisy and imperfect, and their latent dynamics can constantly change and adjust. State-of-the-art methods LILI \cite{xie2020learning} and SILI \cite{wang2021influencing} both acknowledge that their learned policies can be too brittle to interact alongside humans. To better understand how our approach works with actual humans, we conducted an in-person study where participants worked with a 7-DoF robot arm (Panda, Franka Emika). Participants collaborated with this robot to build towers (\fig{front}).

\p{Independent Variables} We compared our approach \textbf{RILI-Transfer} (\fig{main}) to the most recent baseline \textbf{SILI} \cite{wang2021influencing}. For both methods the robot was trained with four simulated latent dynamics prior to the experiment. We hand designed these latent dynamics using strategies that we thought were intuitive for users: building the tower bottom up, top down, or two variations of middle out. In practice, however, we recognize that real human partners could choose one of these dynamics, or come up with some other dynamics entirely.

\p{Dependent Measures} To measure how well the robot coordinated with and influenced the human, we recorded the reward the robot received after each interaction. The robot wanted to build a specific tower: importantly, participants were \textit{never told} what tower the robot wanted. The robot received reward for each block that was placed in the correct order. The robot received a maximum reward of zero if all blocks were in the correct order, and a minimum reward of $-800$ if none of the blocks matched the desired order.

\p{Participants} We recruited $18$ participants ($4$ female, ages $25 \pm 3.3$ years) from the Virginia Tech community. All participants provided informed written consent consistent with university guidelines (IRB $\#20$-$755$). We conducted a between-subject design. Groups varied along two axes: the robot algorithm (\textbf{SILI} or \textbf{RILI-Transfer}) and the human's latent dynamics (explained below). Each user interacted with the robot for $35$ interactions --- the number of interactions was the same for both \textbf{SILI} and \textbf{RILI-Transfer}. 

\p{Procedure} We placed four different colored blocks in the robot's and human's shared workspace: during each interaction the robot pushed these blocks across the table so that they were closer to or farther from the participant. The robot's actions space was the distance it moved each block towards the participant. After the robot finished arranging the blocks the human reached across the table to pick them up and assemble their tower. Once the user finished building the tower, the robot recorded the final tower, observed its reward, and reset the blocks.

Based on our preliminary trials we were concerned that all participants might follow roughly similar latent dynamics. More specifically, we thought all participants might build from the bottom up, putting the closest block on the bottom of their tower and the farthest block at the top. We therefore encouraged participants to follow three different latent dynamics. One group of $6$ participants were instructed to build from the top down; a second group of $6$ were instructed to build from the middle out; and the final $6$ participants were instructed to build from the ends in. The robot had trained with partners that built from the top down and middle out (\textit{Trained Dynamics}), but the robot had never interacted with a partner that built from the ends in (\textit{New Dynamics}). We note that participants did not exactly follow these instructions: if the blocks were close together, participants made random or personalized decisions on how to assemble the tower.

\p{Hypothesis} \textit{Robots that use \textbf{RILI-Transfer} will rapidly learn to influence humans with different latent dynamics.}

\p{Results} The results of our user study are summarized in \fig{exp}. We measured the robot's reward over repeated interactions with each group of users. With \textbf{RILI-Transfer} the reward converged to higher values for all three latent dynamics, while with \textbf{SILI} the robot received fluctuating rewards. To better understand these results we conducted a one-way repeated measures ANOVA with a Sphericity Assumed correction. We found that there was a statistically significant difference in the average cost during the last five interactions ($F(1, 88) = 72.06$, $p<.001$). Examining the robot's behaviors, we noticed that \textbf{SILI} often placed all four blocks at a similar distance from the human --- this caused participants to build random towers, and did not influence them towards the robot's target (see supplementary video). By contrast, \textbf{RILI-Transfer} spaced out the blocks across the table, encouraging the human to take the closest block first and build the desired tower. Overall, these results support our hypothesis and suggest that \textbf{RILI-Transfer} enables robots to learn to influence actual human partners.

%% file: conclusions.tex
\section{Conclusion}

We introduced RILI, an approach that enables robots to robustly influence humans. RILI learns the latent dynamics of changing partners, and identifies a policy that influences each of these partners to coordinate with the ego agent. Our simulations and user study suggest that this approach works with humans who have latent dynamics the robot has not seen before. \textbf{Limitations.} Our approach transfers behaviors learned from previous partners to influence new partners. However, this may fail if the new partner's latent dynamics are sufficiently different from previous partners. For instance, if the robot only trains with humans that build towers from the top down, it may not know any behaviors to coordinate with partners that build from the bottom up.

%% file: main.bbl
\begin{thebibliography}{10}
\providecommand{\url}[1]{#1}
\csname url@rmstyle\endcsname
\providecommand{\newblock}{\relax}
\providecommand{\bibinfo}[2]{#2}
\providecommand\BIBentrySTDinterwordspacing{\spaceskip=0pt\relax}
\providecommand\BIBentryALTinterwordstretchfactor{4}
\providecommand\BIBentryALTinterwordspacing{\spaceskip=\fontdimen2\font plus
\BIBentryALTinterwordstretchfactor\fontdimen3\font minus
  \fontdimen4\font\relax}
\providecommand\BIBforeignlanguage[2]{{%
\expandafter\ifx\csname l@#1\endcsname\relax
\typeout{** WARNING: IEEEtran.bst: No hyphenation pattern has been}%
\typeout{** loaded for the language `#1'. Using the pattern for}%
\typeout{** the default language instead.}%
\else
\language=\csname l@#1\endcsname
\fi
#2}}

\bibitem{habibian2022encouraging}
S.~Habibian and D.~P. Losey, ``Encouraging human interaction with robot teams:
  Legible and fair subtask allocations,'' \emph{IEEE Robotics and Automation
  Letters}, vol.~7, no.~3, pp. 6685--6692, 2022.

\bibitem{hernandez2017survey}
P.~Hernandez-Leal, M.~Kaisers, T.~Baarslag, and E.~M. de~Cote, ``A survey of
  learning in multiagent environments: {D}ealing with non-stationarity,''
  \emph{arXiv preprint arXiv:1707.09183}, 2017.

\bibitem{foerster2018counterfactual}
J.~Foerster, G.~Farquhar, T.~Afouras, N.~Nardelli, and S.~Whiteson,
  ``Counterfactual multi-agent policy gradients,'' in \emph{AAAI}, 2018.

\bibitem{bowling2002multiagent}
M.~Bowling and M.~Veloso, ``Multiagent learning using a variable learning
  rate,'' \emph{Artificial Intelligence}, pp. 215--250, 2002.

\bibitem{cao2018emergent}
K.~Cao, A.~Lazaridou, M.~Lanctot, J.~Z. Leibo, K.~Tuyls, and S.~Clark,
  ``Emergent communication through negotiation,'' in \emph{International
  Conference on Learning Representations}, 2018.

\bibitem{tucker2020adversarially}
M.~Tucker, Y.~Zhou, and J.~Shah, ``Adversarially guided self-play for adopting
  social conventions,'' \emph{arXiv preprint arXiv:2001.05994}, 2020.

\bibitem{losey2019robots}
D.~P. Losey and D.~Sadigh, ``Robots that take advantage of human trust,'' in
  \emph{IEEE/RSJ International Conference on Intelligent Robots and Systems},
  2019, pp. 7001--7008.

\bibitem{foerster2017learning}
J.~N. Foerster, R.~Y. Chen, M.~Al-Shedivat, S.~Whiteson, P.~Abbeel, and
  I.~Mordatch, ``Learning with opponent-learning awareness,'' in \emph{Int.
  Conf. Autonomous Agents and MultiAgent Systems}, 2017.

\bibitem{xie2020learning}
A.~Xie, D.~P. Losey, R.~Tolsma, C.~Finn, and D.~Sadigh, ``Learning latent
  representations to influence multi-agent interaction,'' in \emph{Conference
  on Robot Learning}, 2020.

\bibitem{wang2021influencing}
W.~Z. Wang, A.~Shih, A.~Xie, and D.~Sadigh, ``Influencing towards stable
  multi-agent interactions,'' in \emph{Conf. on Robot Learning}, 2021.

\bibitem{ndousse2021emergent}
K.~K. Ndousse, D.~Eck, S.~Levine, and N.~Jaques, ``Emergent social learning via
  multi-agent reinforcement learning,'' in \emph{International Conference on
  Machine Learning}, 2021, pp. 7991--8004.

\bibitem{sadigh2016planning}
D.~Sadigh, S.~Sastry, S.~A. Seshia, and A.~D. Dragan, ``Planning for autonomous
  cars that leverage effects on human actions,'' in \emph{Robotics: Science and
  Systems}, vol.~2, 2016, pp. 1--9.

\bibitem{saunderson2019robots}
S.~Saunderson and G.~Nejat, ``How robots influence humans: {A} survey of
  nonverbal communication in social human--robot interaction,''
  \emph{International Journal of Social Robotics}, pp. 575--608, 2019.

\bibitem{li2021influencing}
M.~Li, M.~Kwon, and D.~Sadigh, ``Influencing leading and following in
  human-robot teams,'' \emph{Autonomous Robots}, vol.~45, pp. 959--978, 2021.

\bibitem{bestick2016implicitly}
A.~Bestick, R.~Bajcsy, and A.~D. Dragan, ``Implicitly assisting humans to
  choose good grasps in robot to human handovers,'' in \emph{International
  Symposium on Experimental Robotics}, 2016, pp. 341--354.

\bibitem{jonnavittula2022communicating}
A.~Jonnavittula and D.~P. Losey, ``Communicating robot conventions through
  shared autonomy,'' in \emph{IEEE International Conference on Robotics and
  Automation}, 2022.

\bibitem{caruana1997multitask}
R.~Caruana, ``Multitask learning,'' \emph{Machine Learning}, 1997.

\bibitem{carroll2019utility}
M.~Carroll, R.~Shah, M.~K. Ho, T.~Griffiths, S.~Seshia, P.~Abbeel, and
  A.~Dragan, ``On the utility of learning about humans for human-{AI}
  coordination,'' in \emph{NeurIPS}, 2019.

\bibitem{shih2021critical}
A.~Shih, A.~Sawhney, J.~Kondic, S.~Ermon, and D.~Sadigh, ``On the critical role
  of conventions in adaptive human-{AI} collaboration,'' in \emph{International
  Conference on Learning Representations}, 2021.

\bibitem{lupu2021trajectory}
A.~Lupu, B.~Cui, H.~Hu, and J.~Foerster, ``Trajectory diversity for zero-shot
  coordination,'' 2021.

\bibitem{doshi2016hidden}
F.~Doshi-Velez and G.~Konidaris, ``Hidden parameter markov decision processes:
  {A} semiparametric regression approach for discovering latent task
  parametrizations,'' in \emph{IJCAI}, 2016, p. 1432.

\bibitem{cho2014properties}
K.~Cho, B.~van Merri{\"e}nboer, D.~Bahdanau, and Y.~Bengio, ``On the properties
  of neural machine translation: Encoder--decoder approaches,'' in
  \emph{Proceedings of Eighth Workshop on Syntax, Semantics and Structure in
  Statistical Translation}, 2014, pp. 103--111.

\bibitem{haarnoja2018soft}
T.~Haarnoja, A.~Zhou, P.~Abbeel, and S.~Levine, ``Soft actor-critic:
  {O}ff-policy maximum entropy deep reinforcement learning with a stochastic
  actor,'' in \emph{International Conference on Machine Learning}, 2018.

\bibitem{vidal2002probabilistic}
R.~Vidal, O.~Shakernia, H.~J. Kim, D.~H. Shim, and S.~Sastry, ``Probabilistic
  pursuit-evasion games: {T}heory, implementation, and experimental
  evaluation,'' \emph{IEEE Transactions on Robotics and Automation}, vol.~18,
  no.~5, pp. 662--669, 2002.

\end{thebibliography}
